\documentclass[conference]{IEEEtran}
\IEEEoverridecommandlockouts
\usepackage{cite}
\usepackage{url}

\usepackage[linesnumbered,ruled]{algorithm2e}
\usepackage{amsmath,amssymb,amsfonts}
\usepackage{algorithmic}
\usepackage{graphicx}
\usepackage{textcomp}
\usepackage{xcolor}
\usepackage{bm}
\usepackage{multirow}
\usepackage{booktabs}
\usepackage[figuresright]{rotating}
\def\BibTeX{{\rm B\kern-.05em{\sc i\kern-.025em b}\kern-.08em
    T\kern-.1667em\lower.7ex\hbox{E}\kern-.125emX}}

\newcommand{\rui}[1] {{\color{yellow}}{\color{blue}{#1}}}

\begin{document}

\title{Accelerating the Genetic Algorithm for Large-scale Traveling Salesman Problems by Cooperative Coevolutionary Pointer Network with Reinforcement Learning\\}

\author{\IEEEauthorblockN{1\textsuperscript{st} Rui Zhong}
\IEEEauthorblockA{\textit{Grad. School of Info. Science and Technology} \\
\textit{Hokkaido University}\\
Sapporo, Japan \\rui.zhong.u5@elms.hokudai.ac.jp}
\and
\IEEEauthorblockN{ 2\textsuperscript{nd} Enzhi Zhang}
\IEEEauthorblockA{\textit{Grad. School of Info. Science and Technology} \\
\textit{Hokkaido University}\\
Sapporo, Japan \\
enzhi.zhang.n6@elms.hokudai.ac.jp}
\and
\IEEEauthorblockN{3\textsuperscript{rd} Masaharu Munetomo}
\IEEEauthorblockA{\textit{Information Initiative Center} \\
\textit{Hokkaido University}\\
Sapporo, Japan \\
munetomo@iic.hokudai.ac.jp}
}
\maketitle

\begin{abstract}
In this paper, we propose a two-stage optimization strategy for solving the Large-scale Traveling Salesman Problems (LSTSPs) named CCPNRL-GA. First, we hypothesize that the participation of a well-performed individual as an elite can accelerate the convergence of optimization. Based on this hypothesis, in the first stage, we cluster the cities and decompose the LSTSPs into multiple subcomponents, and each subcomponent is optimized with a reusable Pointer Network (PtrNet). After subcomponents optimization, we combine all sub-tours to form a valid solution, this solution joins the second stage of optimization with GA. We validate the performance of our proposal on 10 LSTSPs and compare it with traditional EAs. Experimental results show that the participation of an elite individual can greatly accelerate the optimization of LSTSPs, and our proposal has broad prospects for dealing with LSTSPs.
\end{abstract}

\begin{IEEEkeywords}
Large-scale Traveling Salesman Problems (LSTSPs), Cooperative Coevolution (CC), Pointer Network (PtrNet), Reinforcement Learning (RL)
\end{IEEEkeywords}

\section{Introduction} \label{sec:1}

As a classic NP-hard combinatorial optimization problem, Traveling Salesman Problem (TSP) has been widely studied in the fields of operations research and theoretical computer science\cite{Larranaga:99}. The simplest instance of TSP can be described as: A salesman tries to find the shortest closed route to visit a set of cities under the conditions that each city is visited exactly once. The salesman assumes the distances between any pair of cities are assumed to be known. In real-world problems, there are also many optimization problems based on TSP, such as arranging school bus paths\cite{Lawler:86}, transportation of farming equipment\cite{Lenstra:16}, and so on.

Up to now, many methods for solving TSP have been published, and one of the important branches is Evolutionary algorithms (EAs), such as Genetic Algorithm (GA)\cite{Larranaga:99}, Memetic Algorithm (MA)\cite{Krasnogor:00}, Firefly Algorithm (FA)\cite{Kumbharana:13}, Particle Swarm Algorithm (PSO)\cite{Shi:07}, Immune Algorithm (IM)\cite{Endoh:98}, etc. These naturally inspired algorithms find better solutions by iteration. In recent years, solving TSP through Reinforcement Learning (RL)\cite{Bello:16,Mazyavkina:20} and Deep Learning (DL)\cite{Xin:21} has become a popular theme. Supported by the development of DL technology, modern methods train powerful deep neural networks to learn complex patterns from the TSP instances generated from some specific distributions. The performances of DL models for solving TSP are constantly improved by these works, which unfortunately are still far worse than the strong traditional heuristic solver and are generally limited to relatively small problem sizes\cite{Vinyals:15}.

The curse of dimensionality\cite{Mario:00} is a huge obstacle to solving large-scale optimization problems (LSOPs), the search space of large-scale TSPs (LSTSPs) explodes as the problem dimension increases, which makes heuristic algorithms that perform well in low dimensions often degrade rapidly in high dimensions. Cooperative Coevolution (CC)\cite{Potter:94} is an efficient and mature framework for solving LSOPs. Based on the divide and conquer, CC decomposes LSOPs into multiple sub-problems and solves them separately. This strategy can exponentially reduce the search space of the problem and accelerate the convergence of optimization.

In continuous LSOPs, the key to the successful implementation of the CC framework is the design of decomposition methods\cite{Omidvar:21}, many studies have been published to detect the interactions between variables through nonlinearity check\cite{Tezuka:04}, monotonicity detection\cite{Munetomo:99}, and so on. However, interactions between variables (cities) cannot be detected by these methods in TSP. Based on the hypothesis that "Close cities have a higher probability of having interactions", many clustering methods are applied to TSP for decomposition\cite{Chen:19, Valenzuela:93}. However, traditional clustering methods consume lots of computational costs, which is not friendly to LSTSPs. In addition, since the interactions between variables (cities) cannot be completely detected, the CC framework cannot find the global optimum for LSTSPs. Therefore, we hypothesize that the participation of elite individuals found in the CC framework can accelerate the optimization of LSTSPs. Based on this hypothesis, in this paper, we design a two-stage optimization algorithm to deal with LSTSPs named CCPNRL-GA: In the first stage, inspired by K-Nearest Neighbor (KNN)\cite{Soucy:01}, we choose the nearest $k-1$ cities for a city and form a subcomponent, the Pointer Network (PtrNet) trained by RL is applied to optimize each subcomponent. In the second stage, we combine the sub-tours and form a valid solution to participate in the traditional GA as an elite individual.

The rest of the paper is organized as follows: In Section \ref{sec:2}, the related works are involved. Section \ref{sec:3} introduces CCPNRL-GA in detail. In Section \ref{sec:4}, we show the experimental results of our proposal and its performance comparisons with other heuristic algorithms. Section \ref{sec:5} discusses the future direction of research. Finally, we conclude our paper in Section \ref{sec:6}.

\section{Related works} \label{sec:2}

\subsection{Traveling salesman problem (TSP)} \label{sec:2.1}
In this work, we concentrate on the $2$-D TSP. Given a list of $N$ city coordinates $\{x_1, x_2, ..., x_N \} \in \mathbb{R}^2$ the object is to find the shortest route that each city is visited exactly once and back to the start city finally. In other words, we wish to find an optimal permutation $\sigma$ over the cities that minimizes the tour length:
\begin{equation}
	\label{eq:1}
	\begin{aligned}
		L(\sigma, \mathbf{X})=\sum_{i=1}^{N} \Vert \mathbf{x}_{\sigma(i)}-\mathbf{x}_{\sigma(i+1)} \Vert^2
	\end{aligned}
\end{equation}
where $\sigma(1) = \sigma(N + 1)$, $\sigma(i) \in \{1, ..., N\}$, $\sigma(i) \neq \sigma(j)$ for any $i \neq j$, and $\mathbf{X}=\{\mathbf{x}_1, \mathbf{x}_2, ..., \mathbf{x}_i\}$
is a matrix consisting of all city coordinates $\mathbf{x}_i$.

\subsection{Cooperative Coevolution (CC)} \label{sec:2.2}
Inspired by divide and conquer, CC decomposes the LSTSPs into multiple subcomponents, described as:
\begin{equation}
	\label{eq:2}
	\begin{aligned}
		\min f(\mathbf{x}_1, \mathbf{x}_2, ...,\mathbf{x}_n) = (\min \limits_{c_1}f_1(...,...), ..., \min \limits_{c_m}f_m(...,...))
	\end{aligned}
\end{equation}
where $\mathbf{x}_i$ denotes the coordinate of city $i$, and $f(\cdot)$ represents the objective function. The $c_j$ means a subcomponent in the CC framework, and $f_j (\cdot)$ stands for the objective function in subcomponent $j$. In a study for solving LSTSPs, the CC deals with the problem by dividing it into a set of smaller and simpler subcomponents and optimizing them separately. In summary, there are three main steps consist of the CC framework.

\textbf{Problem decomposition}: Decomposing the LSTSPs into multiple smaller non-overlapping subcomponents. 

\textbf{Subcomponent optimization}: Optimization techniques are applied to each subcomponent.

\textbf{Cooperative combination}: Combining the solutions of all subcomponents to construct the $n$-dimensional solution.

\subsection{Pointer Network with Reinforcement Learning for TSP} \label{sec:2.3}
\begin{figure*}[htb]
	\centering
	\includegraphics[width=12cm]{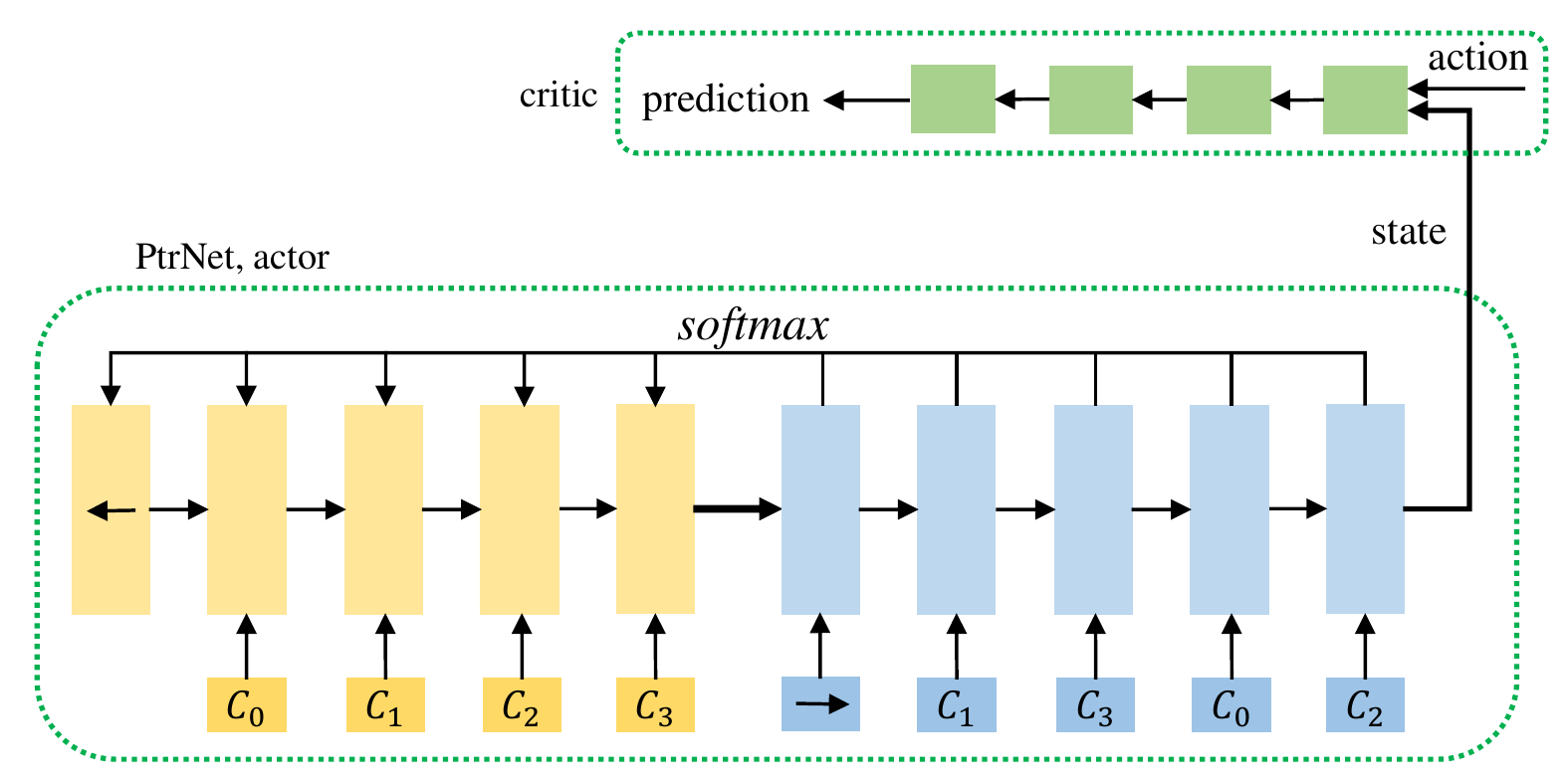}
	\caption{the architecture of PtrNet with RL}
	\label{fig:2}
\end{figure*}

Pointer Network (PtrNet)\cite{Vinyals:15} was proposed by Vinyals et al., which employs a supervised loss function comprising conditional log-likelihood to train the network, and solves the variable-length combinatorial optimization problems flexibly. In paper\cite{Bello:16}, Bello et al. stated that RL provides an appropriate paradigm to train PtrNet for combinatorial optimization problems, especially because these problems have relatively simple reward mechanisms. Fig. \ref{fig:2} shows the architecture of PtrNet with RL.

In Fig. \ref{fig:2}, the PtrNet comprises two Recurrent Neural Network (RNN) modules, encoder and decoder, both of which consist of Long Short-Term Memory (LSTM) cells, and critic comprises three neural network modules: 1) an LSTM encoder, 2) an LSTM process block and 3) a 2-layer ReLU neural network decoder. Model-free policy-based RL (actor-critic) is applied to optimize the parameters of a PtrNet denotes $\bm{\theta}$. The training objective is the expected tour length. Given an input graph $s$, is defined as:
\begin{equation}
	\label{eq:3}
	\begin{aligned}
		J(\bm{\theta}\lvert s)=\mathbb{E}_{\sigma \sim p_\theta(. \lvert s)}L(\sigma \lvert s)
	\end{aligned}
\end{equation}

Paper\cite{Bello:16} adopts policy gradient methods and stochastic gradient descent to optimize the parameters. The gradient of Eq (\ref{eq:4}) is formulated using the well-known REINFORCE algorithm.
\begin{equation}
	\label{eq:4}
	\begin{aligned}
		\nabla_\theta J(\bm{\theta}\lvert s)=\mathbb{E}_{\sigma \sim p_\theta(. \lvert s)}[(L(\sigma \lvert s)-b(s))\nabla_\theta\log p_\theta(\sigma \lvert s)]
	\end{aligned}
\end{equation}
where $b(s)$ denotes a baseline function that does not depend on $\sigma$ and estimates the expected tour length to reduce the variance of the gradients. To estimate the expected tour length, an auxiliary network, $critic$ is introduced, which is parametrized by $\theta_v$, to learn the expected tour length found by our current policy $p_\theta$ given an input sequence $s$. The critic is trained with stochastic gradient descent on a mean squared error objective between its predictions $b_{\theta_v} (s)$ and the actual tour lengths sampled by the most recent policy. The additional objective is formulated as:
\begin{equation}
	\label{eq:5}
	\begin{aligned}
		\mathcal{L}(\theta_v)=\frac{1}{B}\sum_{i=1}^{B} \lVert b_{\theta_v} (s_i) - L(\sigma_i \lvert s_i) \rVert_2^2
	\end{aligned}
\end{equation}
$B$ is the batch size.

\section{CCPNRL-GA} \label{sec:3}
\begin{figure}[htb]
	\centering
	\includegraphics[width=5cm]{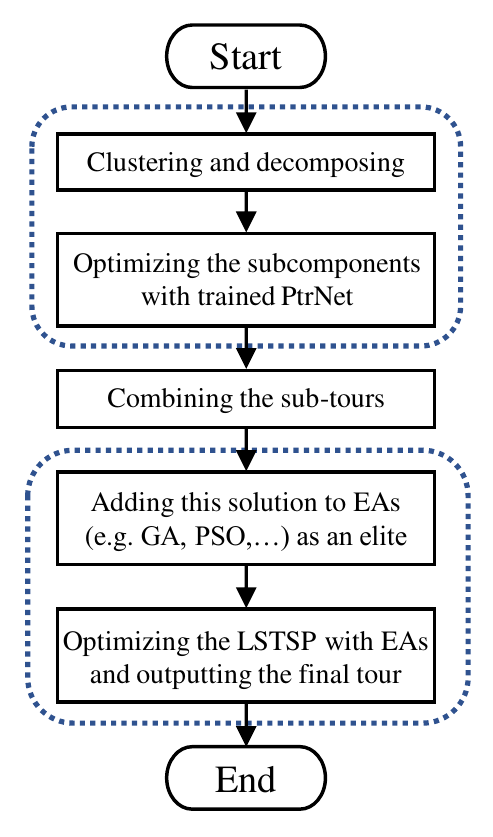}
	\caption{The flowchart of CCPNRL-GA}
	\label{fig:3}
\end{figure}

In this Section, we will describe the main implementation of our proposal in detail. Here in Fig \ref{fig:3}, we demonstrate the flowchart of the main steps. There are two optimization stages in CCPNRL-GA. In the first stage, CC decomposes LSTSPs into many subcomponents, and trained PtrNet is applied to optimize subcomponents. In the second stage, sub-tours found in the first stage are combined and participate in the optimization of GA.

\subsection{A variant KNN}
In the first stage of optimization, we cluster the cities to reduce the search space of LSTSPs by a variant KNN. Inspired by the naive KNN, the closer cities have stronger interactions. Thus, we traverse all cities to calculate the Euclidean distance and form the subcomponents. For each city in the space $C$, if city $i$ is not currently divided into any subcomponent, then we find the nearest $k-1$ cities in the unassigned candidate cities to form a subcomponent. This variant KNN is presented in Eq (\ref{eq:6}).
\begin{equation}
	\label{eq:6}
	\begin{aligned}
	C_k \in C \ {\rm and} \ {\rm unassigned}: \ \arg \min_{k-1}(C_i, C_k)
	\end{aligned}
\end{equation}

\begin{figure*}[htb]
	\centering
	\includegraphics[width=18cm]{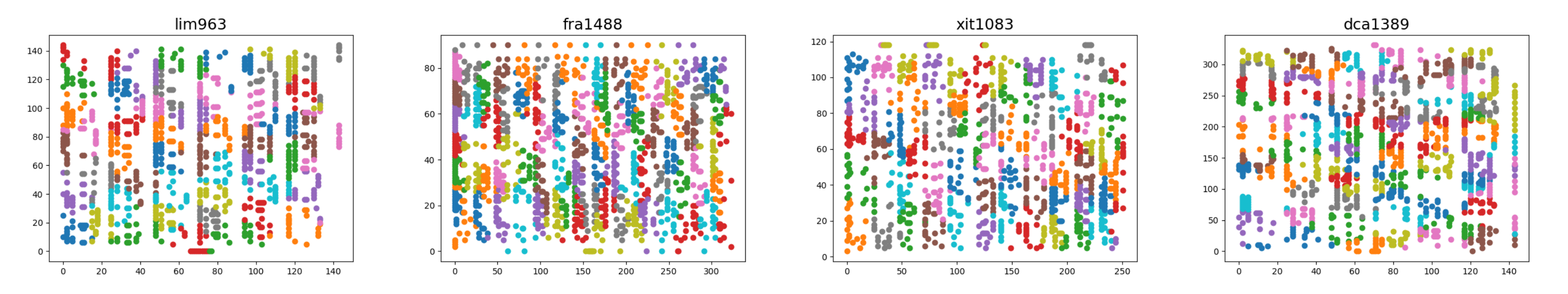}
	\caption{The clustering results in instances of LSTSP ($k=20$).}
	\label{fig:4}
\end{figure*}

Variant KNN divides the cities without iterate correction, which makes it sensitive to the order of city access and inferior to some traditional clustering algorithms in accuracy, such as Kmeans\cite{Bock:07}, ISODATA\cite{Ball:65}, both of them adjust the centroid by distance calculation and iteration, but Variant KNN can save lots of computational costs\cite{Soucy:01}, which is friendly to LSTSPs and can be simply extended to higher dimensional problems. Fig. \ref{fig:4} shows the clustering results in part of instances of LSTSP, the scale of subcomponent $k=20$. We can clearly see that the close cities approximately form a cluster.

\subsection{Optimization with PtrNet}
\begin{figure*}[htb]
	\centering
	\includegraphics[width=18cm]{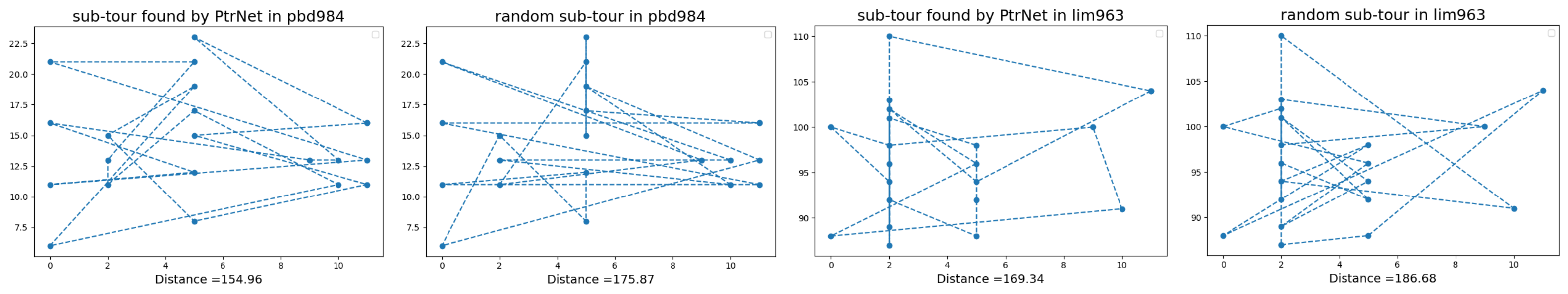}
	\caption{Sub-tours found by random generation and PtrNet optimization in subcomponents}
	\label{fig:5}
\end{figure*}

In the first stage of optimization, we applied the trained PtrNet to optimize each subcomponent, Fig. \ref{fig:5} shows the optimization results of some subcomponents compared with the randomly generated sub-tours. Although the global optima are not achieved, some superior connections are determined by PtrNet, which contribute to the genome with epistasis and participate in the optimization in the second stage.

After all the subcomponents are optimized, we simply disconnect the final connection which is back to the start city, and combine the whole sub-tours to form a valid solution. In the second stage of optimization, this solution is instead of an individual in the random population of GA, which participates in the optimization as an elite. In summary, the pseudocode of CCPNRL-GA is shown in Algorithm \ref{alg:1} 
\begin{algorithm}
	\label{alg:1}
	\caption{CCPNRL-GA}
	\DontPrintSemicolon
	\SetAlgoLined
	\KwIn {${\rm cities}:c;{\rm size \ of\ subcomponents}:k;{\rm Population \ size}:P; {\rm Maximum \ iteration}:M;$}
	\KwOut {${\rm optimal \ tour}: T$}
	\SetKwFunction{FP}{\textbf{CCPNRL-GA}}
	\SetKwProg{Fn}{Function}{:}{}
	\Fn{\FP{$c, k$}}{
	    $S \gets \textbf{QC}(c, k)$ \;
	    $m \gets \textbf{size}(S)$ \;
	    \For{$i=0 \ to \ m$}{
	        $\blacktriangleright$ (Optimize the subcomponents with trained PtrNet) \;
	        $t_i \gets \textbf{PtrNet}(S_i)$ \;
	    }
	    $\blacktriangleright$ (Combine the sub-tours and construct a valid tour) \;
	    $E \gets \textbf{Combine}(t)$ \;
	    $\blacktriangleright$ (Combined tour participates in GA as an elite) \;
	    $T \gets \textbf{GA}(E, P, M)$ \;
		$\textbf{return} \ T$
	}
\end{algorithm}

\section{Numerical Experiments} \label{sec:4}
In this Section, A set of numerical experiments are performed for the algorithm investigations, performance analysis, and comparisons to better understand the performance of our proposal.

\subsection{Experiment Settings} \label{sec:4.1}
\subsubsection{Benchmarks}  \label{sec:4.1.1}

We conduct our experiments on 10 LSTSP instances, Table \ref{tbl:1} shows the details of benchmarks.
\begin{table}[tbh]
	\scriptsize
	\centering
	\caption{The benchmark of 10 LSTSP instances}
	\label{tbl:1}
	\begin{tabular}{ccc}
		\toprule
		Benchmark & number of cities & optimal tour length \\
		\midrule
		dkg813 & 813 & 3199 \\
		lim963 & 963 & 2789 \\
		pbd984 & 984 & 2797 \\
		xit1083 & 1083 & 3558 \\
		dka1376 & 1376 & 4666 \\
		dca1389 & 1389 & 5085 \\
		dja1436 & 1436 & 5257 \\
		icw1483 & 1483 & 4416 \\
		fra1488 & 1488 & 4264 \\
		rbv1583 & 1583 & 5387 \\
		\bottomrule
	\end{tabular}
\end{table}
All benchmarks are downloaded from \cite{Cook:22}

\subsubsection{Compareing methods and parameters} \label{sec:4.1.2}
We compare our proposal with Genetic Algorithm (GA), Particle Swarm Algorithm (PSO), and Immune Algorithm (IA) to evaluate the performance of our proposal, each method is implemented with 30 independent trial runs. the common parameters population size, and maximum iteration are set to 100, and 500 respectively. 

\subsubsection{The details of PtrNet} \label{sec:4.1.3}
\begin{figure}[htb]
	\centering
	\includegraphics[width=9cm]{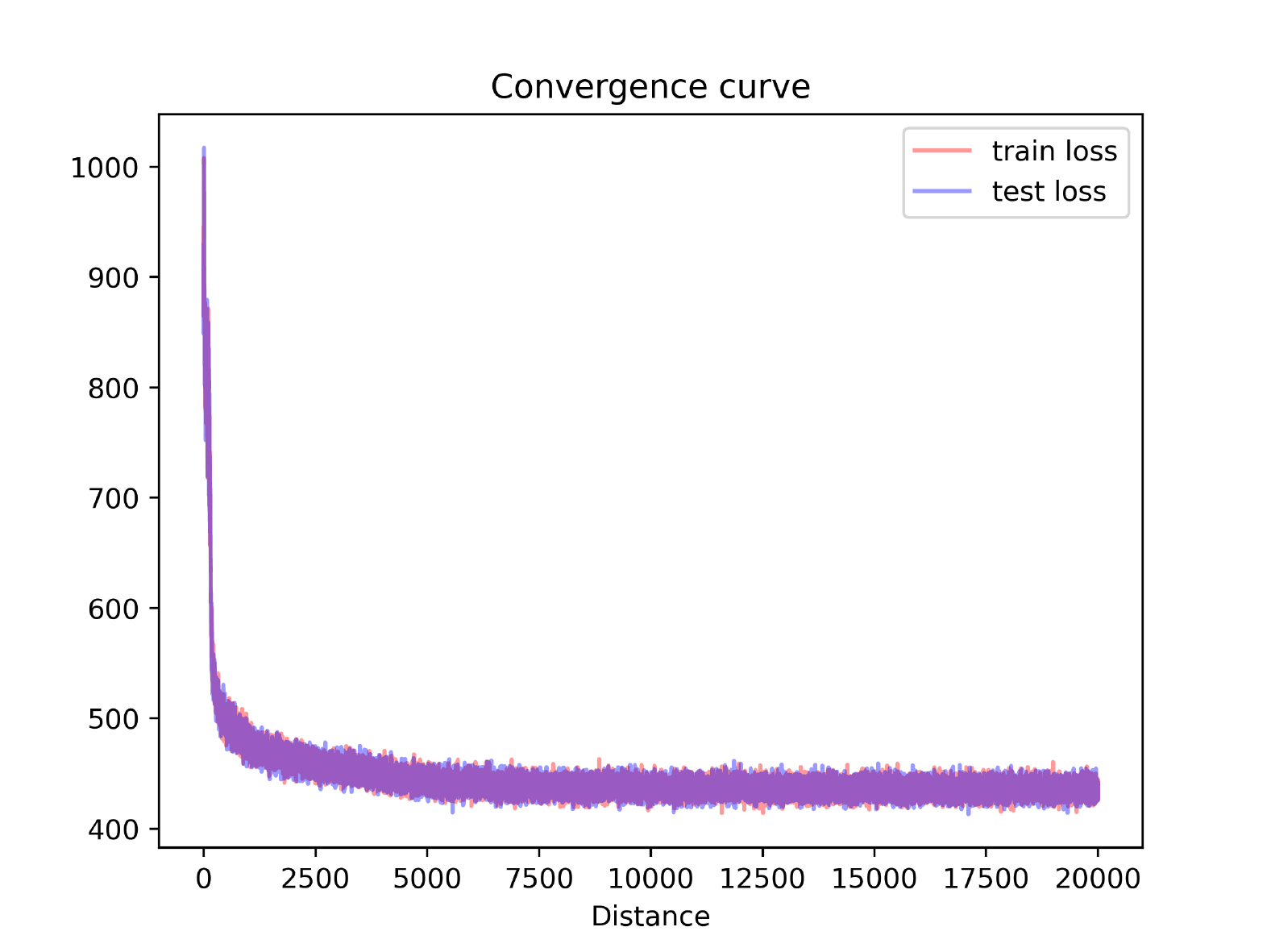}
	\caption{The convergence curve of training PtrNet}
	\label{fig:6}
\end{figure}

We use a mini-batch size of 64, LSTM cells with 128 hidden units, and embed the two coordinates of each point in a 128-D space. Adam optimizer is applied to train our model, and the initial learning rate is 0.001 for each $20$-D subcomponent that we decay every 5000 steps by a factor of 0.96. the maximum iteration of training is 20000. this model is provided by \cite{Yu:21}. The convergence curve of training is shown in Fig. \ref{fig:6}

\subsection{Experimental results} \label{sec:4.2}
\begin{figure}[htb]
	\centering
	\includegraphics[width=8cm]{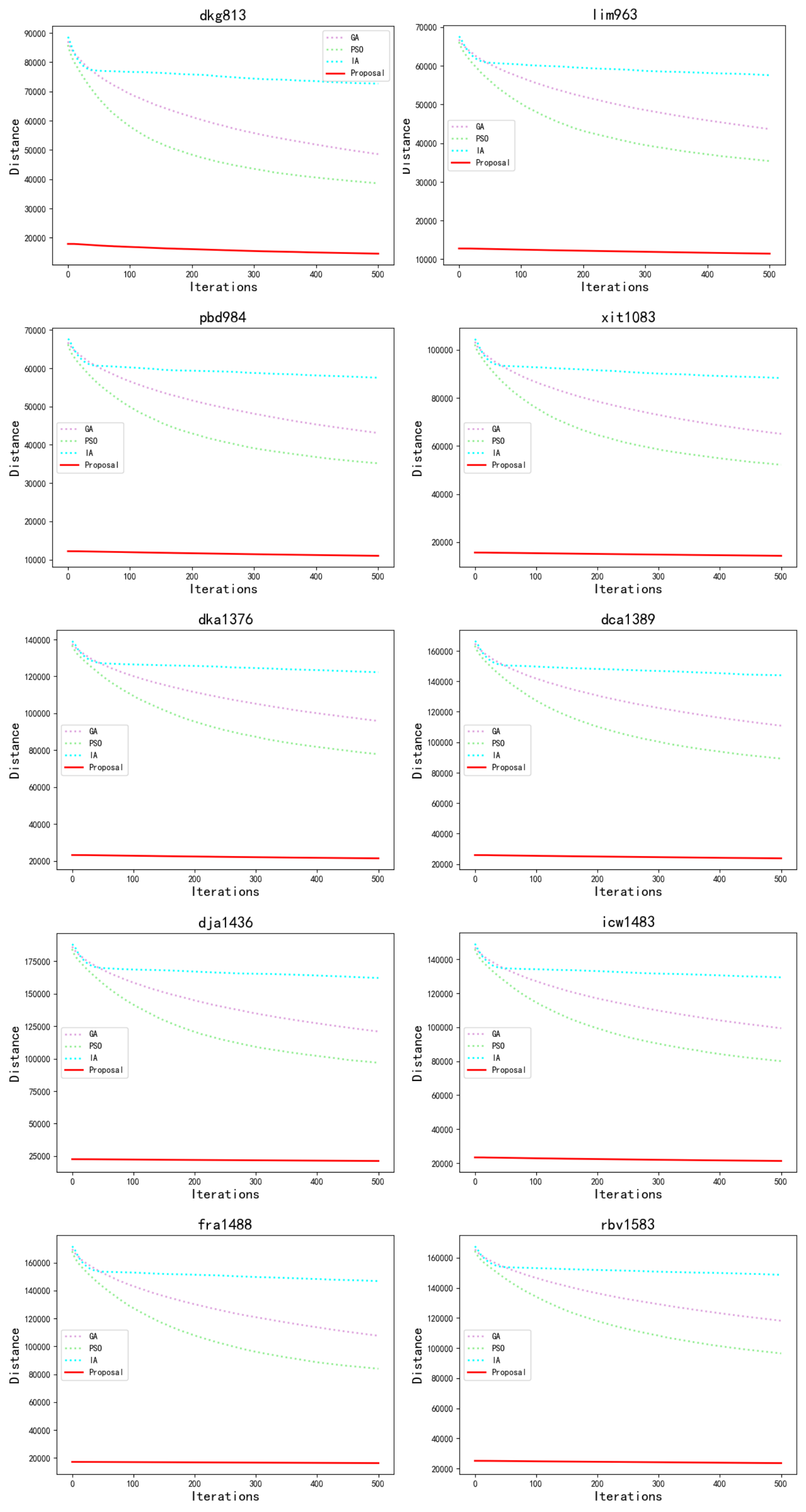}
	\caption{The convergence curves of our proposal and comparison methods}
	\label{fig:7}
\end{figure}

\begin{figure*}[htb]
	\centering
	\includegraphics[width=14cm]{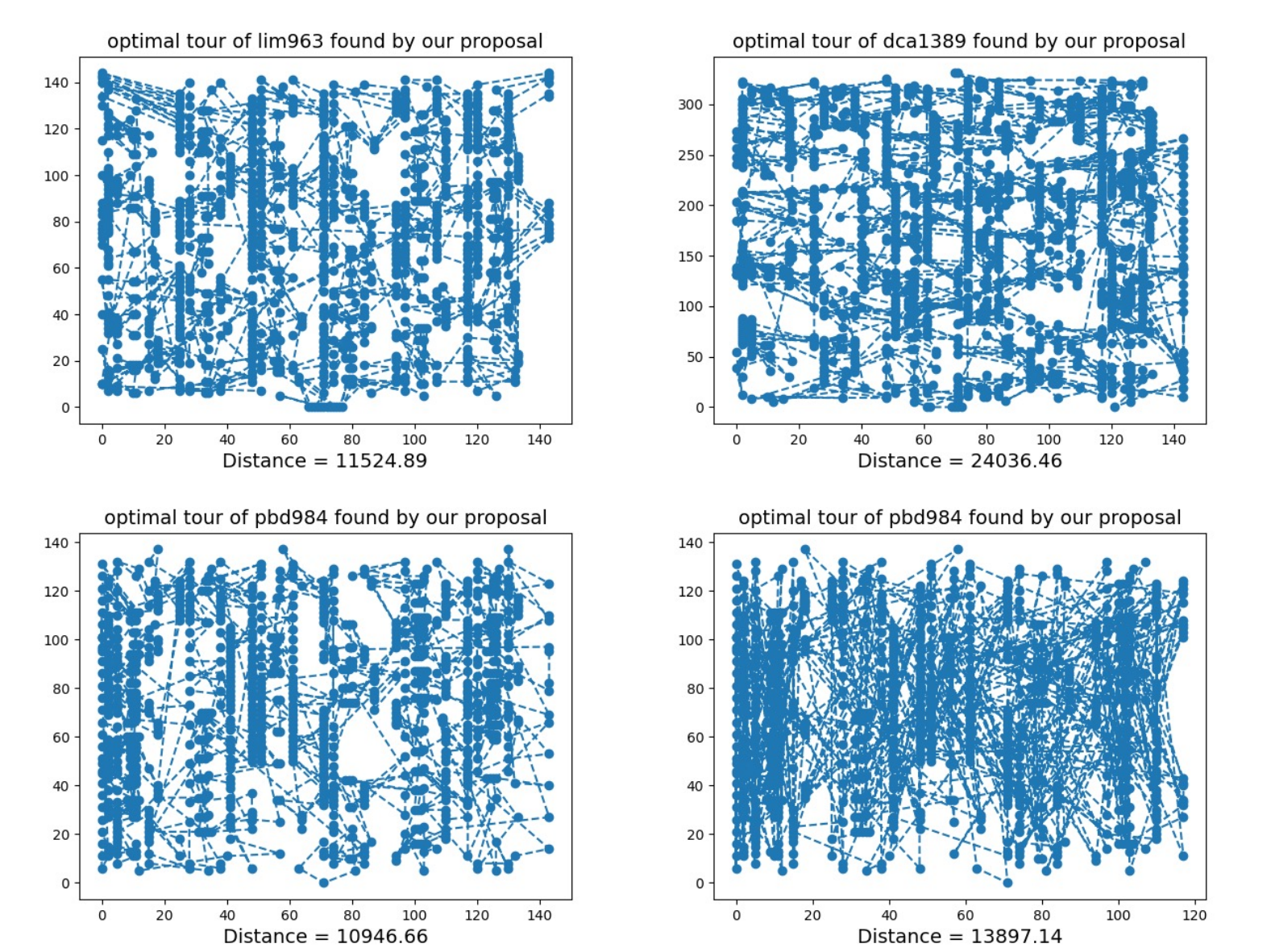}
	\caption{Optimal tour found by our proposal on part of LSTSP instances}
	\label{fig:8}
\end{figure*}

In this section, the performance of our proposal is studied. Experiments are conducted on the benchmark functions presented in Section \ref{sec:4.1.1}. Table \ref{tbl:2} shows the mean and optimal tour length of 30 independent trial runs within our proposal, GA, PSO, and IA. The best solution is in bold. The convergence curves are provided in Fig.\ref{fig:7}. Fig. \ref{fig:8} offers some optimal solutions found by our proposal.

\begin{table*}[tbh]
	\scriptsize
	\centering
	\caption{The mean and optimal tour of our proposal, GA, PSO, and IA in 30 trial runs.  \\
	$M$=Mutation rate, $w$=Inertia weight, $c_1,c_2$=Acceleration coefficients, $T$=Affinity threshold, $\alpha$=Importance of antibody concentration}
	\label{tbl:2}
	\begin{tabular}{ccccccccc}
		\toprule
		\multirow{2}{*}{Benchmark} & \multicolumn{2}{c}{CCPNRL-GA} & \multicolumn{2}{c}{GA$(M=0.01)$} & \multicolumn{2}{c}{PSO$(w=0.8, c_1=c_2=0.1)$} & \multicolumn{2}{c}{IA$(M=0.01, T=0.7, \alpha=0.95)$} \\
		\cmidrule(r){2-3} \cmidrule(r){4-5} \cmidrule(r){6-7} \cmidrule(r){8-9} 
		~ & mean & optimal & mean & optimal & mean & optimal & mean & optimal \\
		\midrule 
		dkg813 & \textbf{14565.30} & \textbf{13619.34} & 48570.88 & 46884.39 & 38613.88 & 36487.93 & 72657.53 & 69659.58 \\
		lim963 & \textbf{11452.69} & \textbf{11059.31} & 43652.80 & 42803.71 & 35381.93 & 33968.81 & 57584.76 & 54083.61 \\
		pbd984 & \textbf{11004.92} & \textbf{10604.12} & 43053.70 & 42378.09 & 35194.06 &  33521.52 & 57501.79 & 54124.48 \\
		xit1083 & \textbf{14294.06} & \textbf{13916.33} & 65028.24 & 63634.57 & 52171.01 & 50410.69 & 88290.96 & 84241.21 \\
		dka1376 & \textbf{21373.52} & \textbf{20721.73} & 95818.51 & 94317.89 & 77817.92 & 75387.33 & 95818.51 & 94317.89 \\
		dca1389 & \textbf{23793.36} & \textbf{23172.27} & 110875.38 & 108744.30 & 89272.90 & 86428.54 & 143935.01 & 138800.28 \\
		dja1436 & \textbf{21277.77} & \textbf{20859.80} & 121004.55 & 119252.48 & 96882.09 & 94177.63 & 162208.18 & 154435.56 \\
		icw1483 & \textbf{21281.81} & \textbf{20694.47} & 99449.41 & 97693.42 & 80080.67 & 77158.91 & 129421.47 & 123916.34 \\
		fra1488 & \textbf{16319.95} & \textbf{16051.10} & 107602.93 & 105103.51 & 83972.63 & 81565.85 & 146794.73 & 137560.83\\
		rbv1583 & \textbf{23547.27} & \textbf{23224.56} & 118178.24 & 116564.78 & 96477.44 & 94071.76 & 148681.26 & 142069.72 \\
		\bottomrule
	\end{tabular}
\end{table*}

\subsection{Analysis} \label{sec:4.3}
The convergence curves in Fig. \ref{fig:7} show that our proposal significantly outperforms other compared methods in initialization, which is due to the participation of an elite individual found in first-stage optimization to second-stage optimization. The objective value of the elite individual is better than the objective value of randomly generated individuals and even better than the optimal solution found by the compared method after 500 rounds of optimization. Experimental results prove that our hypothesis in Section \ref{sec:1} is correct. Furthermore, from the experimental results, an important aspect of solving LSTSPs depending on population-based EAs is the initialization of the population. Population with prior knowledge can greatly accelerate the convergence of optimization.

Meanwhile, we notice that although the optimum found by our proposal still has a difference from the real optimal solution, our proposal tends to converge in the early stage of optimization. We speculate that this is due to the existence of the elite individual, which makes the algorithm fall into a local optimum. How to generate the higher-quality elite individual is one direction of our future research.

\section{Discussion} \label{sec:5}
The above analysis shows our proposal has broad prospects to solve LSTSPs, however, there are still many aspects for improvement. Here, we list a few open topics for potential and future research.

\subsection{More powerful optimizer} \label{sec:5.1}
Although the participation of an elite accelerates the convergence of optimization, there is still a gap between our proposal and the best solution, which is mainly due to the limitation of the optimization algorithm. In future research, we will focus on developing more efficient optimizers both in EAs and Deep Learning fields. One interesting research topic is to introduce the Transformer to solve combinatorial optimization problems.

\subsection{The combination of sub-tours} \label{sec:5.2}
In this experiment, we simply connect the sub-tours in order. Actually, it is worthy and promising to optimize the connection for sub-tour pairs. For every pair of sub-tour which consists of 20 cities, the possible combination is $20\times 20=400$. In future research, a hierarchical optimization for connection is an interesting topic, which means the optimization is executed when a pair of sub-tours is combined, recursively.

\subsection{Dealing with Very Large-scale TSP} \label{sec:5.3}
In this paper, we use LSTSP instances with around 1000 cities as benchmarks. As far as we know, world TSP\cite{William:22} includes more than 1.9 million cities, which is known as the Very Large-scale TSP (VLSTSP). Since all possible combinations of the TSP are $n!$, it is very difficult to solve such a VLSTSP due to the curse of dimensionality. A promising research direction is to employ a certain strategy to generate a population containing prior knowledge to replace random initialization when population-based EAs are applied.

\section{Conclusion} \label{sec:6}
In this paper, we propose a two-stage optimization strategy to deal with LSTSPs. First, we hypothesize that the participation of an elite can accelerate the convergence of optimization for GA, to prove this hypothesis, we construct an elite solution based on the CC framework through PtrNet and RL. Experimental results verify our proposal.  At the end of this paper, we list some interesting topics which can improve our algorithm. Finally, our proposal is a promising study for addressing LSTSPs.

\section{Acknowledgement}
This work was supported by JSPS KAKENHI Grant Number JP20K11967.

\bibliographystyle{IEEEtran}
\bibliography{paper}

\end{document}